\documentclass{article}
\pdfoutput=1
\usepackage{amsmath,amsfonts,bm}

\newcommand{\newterm}[1]{{\bf #1}}

\def\eg{{\textit{e.g.}}}
\def\wrt{{w.r.t. }}

\def\Figref#1{Figure~\ref{#1}}

\def\eqref#1{equation~\ref{#1}}
\def\Eqref#1{Equation~\ref{#1}}
\def\1{\bm{1}}

\def\eps{{\epsilon}}

\DeclareMathAlphabet{\mathsfit}{\encodingdefault}{\sfdefault}{m}{sl}
\SetMathAlphabet{\mathsfit}{bold}{\encodingdefault}{\sfdefault}{bx}{n}

\def\gD{{\mathcal{D}}}

\def\gF{{\mathcal{F}}}

\def\gH{{\mathcal{H}}}

\def\gL{{\mathcal{L}}}

\def\gU{{\mathcal{U}}}
\def\gV{{\mathcal{V}}}

\def\gX{{\mathcal{X}}}

\def\sR{{\mathbb{R}}}

\newcommand{\E}{\mathbb{E}}

\DeclareMathOperator*{\argmax}{arg\,max}
\DeclareMathOperator*{\argmin}{arg\,min}

\usepackage{algorithm}
\usepackage{algorithmic}
\usepackage{natbib}
\usepackage{microtype}
\usepackage{subfigure}

\usepackage{hyperref}

\usepackage{arxiv}

\usepackage{amsmath}
\usepackage{amssymb}
\usepackage{mathtools}
\usepackage{amsthm}

\usepackage{bbm}
\usepackage{url}
\usepackage{hyperref}

\title{Simple Linear Neuron Boosting}

\author{Daniel Munoz \\
        Independent Researcher \\
	\texttt{lnb@dmunoz.org}
}

\begin{document}
\maketitle

\begin{abstract}
Given a differentiable network architecture and loss function, we revisit optimizing the network's
neurons in function space using Boosted Backpropagation \citep{grub2010}, in contrast to optimizing
in parameter space. From this perspective, we reduce descent in the space of linear functions that optimizes
the network's backpropagated-errors to a preconditioned gradient descent algorithm. We show that this
preconditioned update rule is equivalent to reparameterizing the network to whiten each neuron's features,
with the benefit that the normalization occurs outside of inference. In practice, we use this equivalence to
construct an online estimator for approximating the preconditioner and we propose an online, matrix-free
learning algorithm with adaptive step sizes. The algorithm is applicable whenever autodifferentiation is
available, including convolutional networks and transformers, and it is simple to implement for both the
local and distributed training settings. We demonstrate fast convergence both in terms of epochs and wall
clock time on a variety of tasks and networks.
\end{abstract}

\section{Introduction}

First-order optimization techniques for training deep networks on large datasets continue
to yield impressive results on a variety of tasks. Typically, these networks are a composition
of linear functions with nonlinear activations and require multiple epochs over a dataset
that is as large as possible. We refer to these linear functions (\emph{without} the nonlinear activations)
as \newterm{neurons}. However, due to the deeply compositional structure of the network,
achieving fast convergence in practice can sometimes be difficult due to the relative scaling
of values between layers, \eg, vanishing and exploding gradients.
Due to these practical challenges there is wide research into techniques
such as adaptive gradient updates \citep{kingma2014, salimans, rmsprop, momentum},
and improved optimizers \citep{amari1998, Schraudolph2002, martens2020}.

One way to avoid the optimization issues related to feature scaling
(\eg, arbitrarily scaling a feature dimension by a large value typically induces a large gradient vector)
is to use non-parametric methods and forgo optimizing in parameter space, such as with 
Gaussian Processes and kernel methods \cite{kernelbook}.
Another alternative is to optimize in the space of functions, where the corresponding inner product
definition is independent of the parameterization \citep{mason2000,friedman2001}.
In the context of training compositional networks, this idea has
been investigated in \citet{grub2010} where they derive how gradient descent in function space can be used
to backpropagate errors to the respective nested functions; this is referred to as
\emph{boosted backpropagation}.
As will be reviewed in \S \ref{sec:bbp},
this approach no longer explicitly parameterizes each function, but instead operates
in the boosting regime and learns a set of strong-learners, each of which is additive
in a series of weak-learners.

In general, the weak-learners can be any function that is differentiable \wrt
its inputs, including a multilayer perception (MLP), for example. This leads to a design question of whether
to construct (a) a shallow network composed of highly nonlinear strong-learners, 
\eg, each strong-learner is a sum of wide MLP weak-learners,
or (b) a deep sequence of linear strong-learners composed with nonlinear activations. 
For the known state-of-the-art performance they can achieve, we study case (b) of learning
the linear neurons of nonlinear deep networks.

After reviewing the requisite definitions and notations in the next section,
in \S \ref{sec:lnb} we reduce boosted backpropagation for linear neurons to 
a covariant gradient descent algorithm with a metric (inner product space) that corresponds to the
column space of the features that are input to each respective neuron. We discuss how this reduction is equivalent to performing
feature whitening in the network, but the boosting perspective motivates a practical
training algorithm and informs regularization and adaptive step sizes.
Our primary contribution is an easy to implement algorithm to
perform linear boosted backpropagation that is applicable to any network architecture compatible with
autodifferentiation and does not require specialized compute kernels.
A secondary contribution is making the equivalence of this interpretation with prior work such as
Natural Neural Networks \citep{nnn}, FOOF \citep{foof}, and LocoProp \cite{locoprop}.
In addition to ease of implementation, we demonstrate in \S \ref{sec:experiments} that the
algorithm is also efficient with popular architectures in practical settings.

\section{Notation and Background}
We denote the derivative of a function $f: \gU \to \gV$
at $v = f(u)$ as the linear map $\frac{\partial f(u)}{\partial u}: \gU \to \gV$,
or equivalently as $\frac{\partial v}{\partial u}$.
In general, $u$ can represent a high dimensional tensor,
such as a feature map or the filter parameters in a convolutional neural network (CNN).
Similarly, we denote the transpose of the derivative as
$\frac{\partial v}{\partial u}^T: \gV \to \gU$.
Throughout this work we'll operate in vector spaces;
when $f$ is a scalar function, we interchangeably denote
the gradient vector at $u$, $\nabla f(u) \in \gU$, as the transpose of its derivative,
$\frac{\partial v}{\partial u}^T  \in \gU$.

\subsection{Preconditioned Gradient Descent}
\label{sec:pcgd}
We denote $\ell_{(x,y)}: \Theta \to \sR$ as the pointwise loss function for a labeled sample,
where $x$ and $y$ are the inputs and labels, respectively, and the 
samples are drawn from a data distribution, $\gD$.
Minimizing the expected loss, 
$\bar{\ell}(\theta) = \E_{(x,y) \sim \gD} [\ell_{(x,y)}(\theta)]$,
via gradient descent leads to the update rule
$\theta \leftarrow \theta - \alpha^{(t)} \nabla \bar{\ell}(\theta)$,
where $\alpha^{(t)}$ is the desired step size at iteration $t$.
Depending on the form of $\ell$, convergence can be improved by preconditioning
the gradient vector by a matrix $P$, \eg, using the inverse of the Hessian of $\bar\ell(\theta)$ when performing
Newton's method.
In general, when the preconditioner can be factored as $P=WW^T$, the update rule
of following the preconditioned gradient vector, $WW^T \nabla \bar{\ell}(\theta)$, is equivalent to
performing gradient descent on $\bar\ell(\acute\theta)$ with the reparameterized model, $\acute\theta = W\theta$.

\subsection{Natural Gradient Descent}
\label{sec:cgrad}
One way to view natural gradient descent \cite{amari1998} is as
a trust-region optimization problem to find a small step, \wrt some norm,
that is aligned in the direction of $\nabla\bar\ell(\theta)$:
\begin{align}
\argmin_{\delta \theta} & ~ \bar{\ell}(\theta) + \nabla \bar{\ell}(\theta) \cdot \delta\theta \nonumber\\
\text{s.t.} & ~ \frac{1}{2} \langle \delta\theta, \delta\theta\rangle_{M_{\theta}}  = \eps, \label{eq:trust}
\end{align}
where $\langle u, v\rangle_{M_{\theta}} > 0$ is the inner product
defined by the (Riemannian) metric $M_{\theta}$ in the local tangent space at $\theta$.
Solving for the stationary point of its Lagrange function results in an update step
$\delta \theta \propto M_{\theta}^{-1} \nabla \bar{\ell}(\theta)$.
When defining $M_\theta$ as the Fisher Information Matrix (FIM), the resulting
vector is referred to as the natural gradient vector; see \citet{kunster2019} for an informative review and discussion.
However, in general, we are free to choose any positive-definite metric, $M_{\theta} \succ 0$,
and we denote the resulting vector under the metric as $\nabla_{M} \ell(\theta) = M_{\theta}^{-1} \nabla \ell(\theta)$.
Note that taking a step solely proportional $\nabla_{M}\bar{\ell}(\theta)$
ignores the trust region constraint of \Eqref{eq:trust};
solving for it obtains the appropriate step size, $\alpha_{\theta}$,
in the original parameter space that induces an  $\eps$-sized step under the chosen metric:
$\alpha_{\theta} = \sqrt{\frac{\eps}{z_\theta}}$, where
\begin{equation}
\label{eq:metricnorm}
z_\theta
= \langle \nabla_{M} \bar{\ell}(\theta), \nabla_{M} \bar{\ell}(\theta) \rangle_{M_{\theta}}
= \nabla_{M} \bar{\ell}(\theta) \cdot \nabla \bar{\ell}(\theta).
\end{equation}

\subsection{Functional Gradient Descent}
\label{sec:fgrad}
In contrast to optimizing a function in the space of parameters,
an alternative is to optimize in the space of functions, $\gF$, which
can also be viewed as boosting \citep{mason2000, friedman2001}.
The following summarizes \S2 of \citet{grubbthesis} to introduce notation on this topic.
We denote $\bar{\gL}[f] = \E_{(x,y) \sim \gD}[l_{y}(f(x))]$ to be the loss functional
that computes the expected loss for a given function $f: \gX \to \gV$,
where $l_y: \gV \to \sR$ is loss function (using label $y$) in the image (output space) of $f$,
and $\gV$ is an application-specific vector
space\footnote{In general, each element in $\gV$ need not be a 1-D vector and each could
be a multi-dimensional tensor. 
For example, in $k$-class logistic regression, the image of $f$ is the predicted logits vector
($\gV = \sR^k$) and $l$ is pointwise log-loss.
Whereas for $k$-class semantic image segmentation, each the image of $f$ is a tensor
($\gV = \sR^{h \times w \times k}$) whose outer component is the per-pixel logits 
and $l$ is the per-pixel log-loss summed over the inner spatial components.}.
Letting $v_x = f(x)$, the functional gradient ``vector'' of $\bar{\gL}$ at $f$ is defined as the
\emph{function},
\begin{equation*}
\nabla_F \bar{\gL}[f] = \E_{(x,y) \sim \gD} [\nabla_F l_y(v_x)] = \E_{(x,y) \sim \gD} [ \lambda_{(x,y)} \mathbbm{1}_x],
\end{equation*}
where each $\lambda_{(x,y)} = \frac{\partial l_y(v_x)}{\partial v_x}^T \in \gV$
is a gradient vector, and $\mathbbm{1}_x: \gX \to \{0, 1\}$ is the Dirac delta function centered at $x$.
For brevity, let $\triangle_f = \nabla_F \tilde{\gL}[f]$.

Instead of the defining the strong-learner, $f$, additively in $\triangle_f \in \gF$,
each $\triangle_f$ is projected onto a smaller hypothesis space, $\gH \subseteq \gF$,
such as small decision trees or MLPs, in order to generalize.
The projection of $\triangle_f$ onto a hypothesis $h \in \gH$ is analogous to vector projection in Euclidean space:
$\frac{\langle \triangle_f, h \rangle_F}{\Vert h \Vert_F}\frac{h}{\Vert h \Vert_F}$,
where $\langle f,g \rangle_F = \mathbb{E}_x[f(x) \cdot g(x)]$ is the inner product in function
space with norm $\Vert h \Vert_F = \sqrt{\langle h,h \rangle_F}$.

When $\gH$ are regressors, the hypothesis, $\hat{h}$, that maximizes the scalar projection term is
equivalent \citep{friedman2001} to minimizing a least squares problem,
\begin{equation}
\label{eq:scalarproj}
\hat{h} = \argmax_{h \in \gH} \frac{\langle \triangle_f, h \rangle_F}{\Vert h \Vert_F}
= \argmin_{h \in \gH} \E_{x} [ \Vert h(x) -\triangle_f(x) \Vert^2 ].
\end{equation}
In practice, this translates to training a (vector-output) regressor
over the dataset $\{(x,\lambda_{(x,y)})\}$.
Finally, this leads to the update rule
$f \leftarrow f - \eps^{(t)} \frac{\hat{h}}{\Vert \hat{h} \Vert_F}$,
and we refer to each $\hat{h}$ as weak-learners.

\subsubsection{Regularization}
\label{sec:reg}
In addition to minimizing the incurred loss of $f$, we may also want to include
a regularization term in the objective $\tilde{\gL}[f] + \frac{\rho}{2} \Vert f \Vert_F^2$, where 
$\rho \geq 0$. The update rule with the regularized objective is
\begin{equation}
\label{eq:reg}
f \gets (1-\eps^{(t)}\rho) f -  \eps^{(t)} \frac{\hat{h}}{\Vert \hat{h} \Vert_F}.
\end{equation}
In practice, this is implemented by shrinking the existing weak-learner coefficients $\{\eps^{(j)}|j<t\}$
by the factor $0 \leq (1-\eps^{(t)}\rho) \leq 1$.

\subsection{Boosted Backpropagation}
\label{sec:bbp}

For simplicity in explanation, we begin with the problem of training a feed-forward
network (FFN), as done in \citet{grub2010}, and note that general architectures
will be discussed in the next section.
We represent a FFN as a composition of $m$ differentiable functions
$F = \{f_i:\gX_{i-1} \to \gX_i| 1 \leq i \leq m \}$, where a subset $F_B \subseteq F$
are neurons with parameters $\theta_i$ that we want to train, \eg,
$f_i(\cdot;\theta_i) \in F_B$ could be convolutional layer with filters $\theta_i$.
We denote $\theta_B$ as all the trainable parameters in the network.
The complement set of functions, $F \setminus F_B$, are fixed, \eg, activation functions, resizing layers, etc.
We denote the loss incurred for sample $(x,y)$ for the given network parameters
as $\ell_{(x,y)}(\theta_B) = (l_y \circ f_m \circ \ldots \circ f_1)(x)$,
where $l_y: \gX_m \to \sR$ computes the loss of the last layer's prediction vector with label $y$.
Backpropagation can be used to compute the gradient vector of the
expected loss, $\nabla \bar\ell(\theta_B)$,
and we denote $g_i$ to be the component of $\nabla \bar\ell(\theta_B)$
that corresponds to $\theta_i$, i.e., $g_i = \frac{\partial\bar\ell(\theta_B)}{\partial \theta_i}^T$.

Forgoing parameterizing the FFN neurons with weights,
\citet{grub2010} optimize the loss \wrt each neuron's \emph{function}
by using the adjoint state method for
the equivalent constrained minimization problem,
\begin{align}
\argmin_{F_B} ~& \E_{(x,y) \sim \gD} [l_y(x_{m})] \nonumber \\
\begin{split}
\label{eq:forward}
\text{s.t.} ~& x_0 = x, x_{i} = f_{i}(x_{i-1}), i \in [1, \ldots, m].
\end{split}
\end{align}
Using the necessary conditions for a stationary point of its Lagrangian,
they describe an algorithm to recursively compute the functional gradient vector for each neuron.
For $f_i \in F_B$ in a FFN, its functional gradient vector is $\lambda_i \mathbbm{1}_{x_{i-1}}$,
where
\begin{align}
\lambda_{m+1} &= \frac{\partial l_y(x_m)}{\partial x_m}^T, \nonumber \\
\lambda_{i} &= \frac{\partial x_i}{\partial x_{i-1}}^T \lambda_{i+1},  i \in [1, \ldots, m],
\label{eq:costates}
\end{align}
are the errors backpropagated from the loss layer via vector-Jacobian products (VJPs).
That it is, each training step is analogous to performing normal backpropagation
with the key difference of training a weak-learner using the dataset 
$\{(x_{i-1}, \lambda_i)\}$ for each $f_i \in F_B$,
instead of pulling-back the targets, $\lambda_i$, through respective neuron's local derivative.
That is, $f_i$'s component of $\nabla \bar\ell(\theta_B)$ is
$g_i = \frac{\partial x_i}{\partial \theta_i}^T \lambda_i$.

As presented, the time complexity for computing the unprojected
functional gradient vectors is the same backpropagation; however,
the storage complexity increases linearly with the dataset and dimensionality
due to aggregating the regression targets.
The projection operation then adds significant compute as it requires solving a
large vector regression problem for each neuron\footnote{\eg,
for a convolutional layer, each $\lambda_i$ represents a
$\sR^{h \times w \times d}$ feature map and there are $|\gD|$ of them.}.
Lastly, the presented algorithm is specific to a FFN and it is left
as an exercise to construct the corresponding recurrence relation of \Eqref{eq:costates}
for different network architectures, which can be error prone.
In the next section we address these concerns and provide a simple and efficient boosting algorithm
for any network architecture compatible with autodifferentiation.

\section{Boosting over Linear Neurons}
\label{sec:lnb}

\subsection{Reduction}
For training general architectures we can forgo computing the 
regression targets, $\lambda_i$. Instead, we use
its definition that it is the gradient vector of the loss
\wrt the \emph{output} of $f_i$,
\begin{equation}
\label{eq:lambda}
\lambda_i^T
= \frac{\partial l_{y}(x_m)}{\partial f_i(x_{i-1})}
= \frac{\partial l_{y}(x_m)}{\partial x_i},
\end{equation}
where $x_{i-1} \in \gX_{i-1}$ are the forward-propagated \emph{input} features to the neuron $f_i$.
Note that $x_{i-1}$ and $\lambda_i$ both depend on the sample $(x,y)$ and current
network definition, $F_B$.

When projecting the functional gradient vector for $f_i$
over a linear hypothesis space, the weak-learners have the same
function definition as the neuron but differ in the parameters.
The scalar projection step (\Eqref{eq:scalarproj}) then corresponds to
solving the ordinary least squares (OLS) problem
\begin{equation}
\label{eq:OLS}
\hat{\theta}_i = \argmin_{\theta} \frac{1}{2} \E_{(x,y) \sim \gD} 
\Vert f_i(x_{i-1};\theta) - \lambda_i \Vert^2.
\end{equation}
Because this is an OLS problem, we know its solution
is the projection of the targets, $\{\lambda_i\}$,
onto the column space of the forward-propagated input features to $f_i$;
we denote this set of input features as $X_i = \{x_{i-1}\}$.

Using \Eqref{eq:lambda} and the definition
$x_i = f_i(x_{i-1};\theta)
= \frac{\partial x_i}{\partial \theta_i}\theta$
(because $f_i$ is linear),
we can rewrite \Eqref{eq:OLS} as
\begin{equation}
\label{eq:OLS2}
\hat{\theta}_i = \argmin_{\theta} \frac{1}{2} \E_{(x,y) \sim \gD}
\biggl \Vert \frac{\partial x_i}{\partial \theta_i}\theta - \frac{\partial l_{y}(x_m)}{\partial x_i}^T \biggr \Vert^2,
\end{equation}
which has the corresponding Normal Equation
\begin{align}
\E_{x_{i-1} \sim X_i} \biggl [ \frac{\partial x_i}{\partial \theta_i}^T\frac{\partial x_i}{\partial \theta_i} \biggr ] \hat{\theta}_i 
= \E_{(x,y) \sim \gD} \biggl [ \biggl (\frac{\partial l_{y}(x_m)}{\partial x_i} \frac{\partial x_i}{\partial \theta_i} \biggr )^T \biggr ].
\label{eq:normal}
\end{align}
The right-hand side is the component of $\nabla \bar\ell(\theta_B)$ corresponding to $\theta_i$, so the solution is
$\hat{\theta}_i = M_i^{-1}g_i$, where
 $M_i = \E_{x_{i-1} \sim X_i} [ \frac{\partial x_i}{\partial \theta_i}^T  \frac{\partial x_i}{\partial \theta_i}]$.
We can confirm that the norm in function space is equivalent to the
inner product under $M_i$ of \Eqref{eq:metricnorm}:
$$
\Vert f_i(\cdot; \hat{\theta}_i) \Vert^2_F
= \E \biggl [\frac{\partial x_i}{\partial \theta_i} \hat{\theta}_i \cdot \frac{\partial x_i}{\partial \theta_i} \hat{\theta}_i \biggr ]
= \hat{\theta}_i \cdot M_i \hat{\theta}_i
= \hat{\theta}_i \cdot g_i.
$$

Because the OLS problems are solved independently per neuron,
the final solution is equivalent to as if we constructed a block-diagonal metric, $M$,
containing each $M_i$ along the diagonal and then computing the preconditioned gradient
vector from \S \ref{sec:cgrad}. We denote this vector as $\hat{\theta}_B = \nabla_M \bar\ell(\theta_B)$,
which is composed of each neuron's $\hat{\theta}_i$ solution.
Note that the structure of the network is only needed to compute $\nabla \bar\ell(\theta_B)$ (via
autodifferentiation) and we can then solve \Eqref{eq:normal} for each neuron independently.

\subsection{Optimization}
To solve for $\hat{\theta}_i$ in \Eqref{eq:normal},
we can leverage any linear system solver that internally
uses matrix-vector-products (MVP) instead of instantiating
each metric, $M_i$.
First, we need to save the set of forward-propagated input features, $X_i$, to the neuron; we note
that these features are already computed and saved in
memory during backpropagation.
In practice, the MVP $M_i \theta$ can then be computed via a
VJP composed with a Jacobian-vector-product (JVP) of
the vectorized function $\mathbf{f}_{X_i}(\theta_i) = \text{vec}(\{f_i(x_{i-1};\theta_i)\}_{x_{i-1} \in X_i})$,
which maps $f_i(\cdot; \theta_i)$ over  every element (tensor) in dataset $X_i$, and then
\begin{equation}
\label{eq:mvp}
M_i(\theta) = 
\frac{1}{n_i} \frac{\partial \mathbf{f}_{X_i}}{\partial \theta_i}^T \frac{\partial \mathbf{f}_{X_i}}{\partial \theta_i} \theta,
\end{equation}
where $n_i$ is the number of samples that contribute to the
gradient\footnote{In general this value can be computed using the shape of $\gX_i$,
\eg, if $f_i(x;\theta) = \theta^Tx$, then $n_i$ is
the batch size; if $f_i$ is a strided convolution, then $n_i$ 
is the number of output pixels in the batch.}.
Note that since $\mathbf{f}_{X_i}$ is linear it need only be
linearized once (at any point) by the linear system solver.

After solving \Eqref{eq:normal} for each neuron, we can construct $\hat{\theta}_B$ and
compute the $\eps$-sized step under the full metric $M$ via \Eqref{eq:metricnorm}.
Lastly, we may wish to regularize our strong-learner in function space as discussed in 
\S \ref{sec:reg}. Because the learner is linear, this corresponds to a 
``weight decay'' of factor $1-\sqrt{\eps} \rho$; note that if a neuron contains
bias terms, these should also decay.
Algorithm \ref{algo1} summarizes the entire optimization algorithm, which
we refer to as Linear Neuron Boosting (LNB).

\begin{algorithm}[tb]
\caption{Linear Neuron Boosting}
\label{algo1}
\begin{algorithmic}[1]
\INPUT{Network $F$,
linear neurons $F_B \subseteq F$ with parameters $\theta_B$,
loss function $\bar\ell$,
step size schedule $\{\eps^{(t)}\}_{t=1}^T$,
weight decay $\rho$}
\STATE Randomly initialize $\theta_B$
\FOR{$\eps \in [\eps^{(1)}, \ldots, \eps^{(T)}]$}
\STATE Compute $\nabla \bar\ell(\theta_B)$ via backpropagation and save the 
inputs to each neuron, $X_B = \{X_i | f_i \in F_B\}$ \label{algo:bp}
   \FOR{$(f_i, g_i, X_i) \in (F_B, \nabla\bar\ell(\theta_B), X_B)$}
      \STATE Define the \Eqref{eq:mvp} MVP function, $M_i(\cdot)$, using the VJP and JVP of $\mathbf{f}_{X_i}$
      \STATE $\hat{\theta}_i \gets \texttt{linear\_solver}(A=M_i, b=g_i)$
   \ENDFOR
   \STATE $\theta_B \gets (1-\sqrt{\eps}\rho)\theta_B$  \hfill $\triangleright$ Weight decay
   \STATE $z = \hat{\theta}_B \cdot \nabla \bar \ell(\theta_B)$  \hfill $\triangleright$ Norm
\STATE $\theta_B \gets \theta_B - \sqrt{\frac{\eps}{z}} \hat{\theta}_B$ \hfill $\triangleright$ Adaptive step size
\ENDFOR
\OUTPUT{$\theta_B$}
\end{algorithmic}
\end{algorithm}

\subsection{Update Properties}
\label{sec:WW}
We now analyze the properties of the LNB update rule. For simplicity in explanation,
we use a neuron with scalar output, $f_i : \gX_{i-1} \to \sR$, and note that this is sufficient because
the OLS solution of \Eqref{eq:OLS} for a vector output is the same as independently solving for each component.

When $f_i(x;\theta) = \theta^Tx$ \emph{without} a bias term, the resulting metric is the \emph{uncentered}
second moment matrix, $M_i = \E_{x \sim X_i}[xx^T]$. Denoting the number of parameters as $d_i$,
when the features are linearly independent and $n_i > d_i$, then $M_i \succ 0$ and the
OLS solution is scale equivariant \wrt the features. Of course, satisfying both of these
conditions can be rare in practice and we discuss mitigations in \S \ref{sec:PSD}.

When $f_i(x;\theta) = \theta^T \acute x$, where $\acute x^T = [x^T, 1]$
the resulting metric has the block form
\begin{equation}
M_i = \begin{pmatrix}
\Sigma_i + \mu_i \mu_i^T & \mu_i\\
\mu_i^T & 1
\end{pmatrix},
\end{equation}
where $\mu_i = \E_{x \sim X_i}[x]$ and $\Sigma_i$ are the mean and covariance matrix of $X_i$, respectively.
We can analytically invert $M_i$ via the factorization $M_i=L_i^T L_i$, where
$L_i = \begin{pmatrix}
   \Sigma_i^{\frac{1}{2}} & 0\\
   \mu_i^T & 1
\end{pmatrix}$.
Then, $M_i^{-1} = W_i W_i^T$, where
$W_i = \begin{pmatrix}
   \Sigma_i^{-\frac{1}{2}} & 0\\
   -\mu_i^T \Sigma_i^{-\frac{1}{2}} & 1
\end{pmatrix}$. From \S \ref{sec:pcgd}, preconditioning $W_iW_i^T g_i$ is the same
as reparameterizing the model:
$$
f_i(x; W_i\theta) = (W_i \theta)^T \acute x = \theta^T W_i^T \acute x = \theta^T (\Sigma_i^{-\frac{1}{2}}(x-\mu_i)).
$$
To summarize, when running LNB with neurons that have bias parameters,
it is equivalent to optimizing a model where we 
added a whitening transformation step in the network to each $f_i$'s input features using the
respective statistics from $X_i$, \emph{before} applying the parameters.
We further discuss this relationship to Batch Normalization \citep{batchnorm} and other methods in \S \ref{sec:connections}.

When the neuron is a convolution layer, $f_i(x;\theta) = x * \theta$, the output is still linear in the parameters (filters)
and the interpretation reduces to either of the previous two cases, with the only difference that the
first and second moments are computed using the filter's spatial support features (and are
translationally equivariant). By representing the metric using Jacobians, we can use autodifferentiation
to compute the metric without requiring a specialized implementation for this case, \eg, $\texttt{unfold}$.

Other common forms of $f_i$ will likely result in an identity $M_i$. For example, in a vision
transformer \citep{vit} the corresponding metric for both the embedding function, $f(x;\theta) = x + \theta$, and the class-token
function, $f(x;\theta) = \texttt{concat}(\theta, x)$, is the identity. In these cases, the linear solver can be skipped
since $\hat{\theta}_i = g_i$ and there are no other changes to the LNB algorithm.

\subsection{Positive Semi-definite Metrics}
\label{sec:PSD}
Since $M_i$ is an inner product of a Jacobian with itself, $M_i \succeq 0$, in general.
To ensure $M_i \succ 0$ in practice, a common regularizer is to add a small constant, $\gamma$,
along the diagonal of $M_i$ (excluding the bias terms) when solving the linear system. 
In the boosting perspective this adds a $\gamma \Vert \theta \Vert^2$ penalty term to \Eqref{eq:OLS2}
(which is why one would typically exclude the bias terms).
Because this regularizes the norm of the parameters, the solution is no longer scale equivariant.
When the metric's eigenvalues are small it no longer informs a meaningful
step-size ($z_{\theta} \approx 0$). In this case, we can upperbound the maximum allowed step size
to avoid taking a large step. This is essentially equivalent to disabling the
preconditioning and following the typical gradient vector direction.

\begin{algorithm}[ht]
   \caption{Online Linear Neuron Boosting}
   \label{algo2}
   \begin{algorithmic}[1]
   \INPUT{Network $F$,
   linear neurons $F_B \subseteq F$ with parameters $\theta_B$,
   mini-batch loss function $\bar\ell$,
   step size schedule $\{\eps^{(t)}\}_{t=1}^T$,
   weight decay $\rho$, minimum norm $z_0$}
   \STATE Randomly initialize $\theta_B$
   \FOR{$\eps \in [\eps^{(1)}, \ldots, \eps^{(T)}]$}
   \STATE Compute $\nabla \bar\ell(\theta_B)$ via backpropagation and save the 
   inputs to each neuron, $X_B = \{X_i | f_i \in F_B\}$
   \STATE $g_B = \texttt{ema}(g_B, \nabla \bar\ell(\theta_B))$
      \FOR{$(f_i, g_i, X_i) \in (F_B, g_B, X_B)$}
         \STATE $\mu_i = \texttt{ema}(\mu_i, \texttt{mean}(X_i))$
         \STATE $\chi_i = \texttt{ema}(\chi_i, \texttt{mean}(X_i \odot X_i))$
         \STATE Create $P_i$ from $\mu_i$ and $\chi_i$ (\S \ref{sec:online})
         \STATE Define the \Eqref{eq:mvp} MVP function, $M_i(\cdot)$, using the VJP and JVP of $\mathbf{f}_{X_i}$
         \STATE $\hat{\theta}_i \gets \texttt{cg}(A=M_i, b=g_i, P=P_i, x_0=\hat{\theta}_i)$
      \ENDFOR
      \STATE $\theta_B \gets (1-\sqrt{\eps}\rho)\theta_B$  \hfill $\triangleright$ Weight decay
      \STATE $z = \max(z_0, \hat{\theta}_B \cdot g_B)$  \hfill $\triangleright$ Norm
   \STATE $\theta_B \gets \theta_B - \sqrt{\frac{\eps}{z}} \hat{\theta}_B$ \hfill $\triangleright$ Adaptive step size
   \ENDFOR
   \OUTPUT{$\theta_B$}
   \end{algorithmic}
\end{algorithm}

\subsection{Online Learning}
\label{sec:online}
As presented, each step of boosting should be performed using the entire dataset,
which is quite impractical. We can easily switch to using an online estimator
of the gradient vector, \eg, an exponential moving average (EMA) over minibatches.
However, computing an online estimate for $M_i$ in our setting is difficult because we
rely on MVPs and do not materialize it nor $M_i^{-1}$.
One alternative, at the expense of increased compute and memory during training,
is to estimate $M_i$ using samples from a much larger dataset, which has demonstrated
to help under the FIM \citep{pascanu-2014}. Because $M_i$ does not require labels
we can compute it via only forward passes.
A second option, when $M_i$ can be materialized in memory, is to use standard techniques for
online covariance estimation \citep{dasgupta}.
If eigendecomposition can be afforded, a third option is to initialize $M_i^{-1}$ with the identity and
perform k-rank updates via the Woodbury matrix identity; this could also be amortized across gradient updates.

Instead of the previous three options, we instead propose to use an online, approximate
estimate of the feature moments as a preconditioner for the linear system solver.
Specifically, we use the Conjugate Gradient algorithm where we precondition each conjugate
step with the approximate solution to facilitate quick convergence.
The form of the preconditioner, $P_i$, is dependent on if the respective $f_i$ has a bias term or not.
If $f_i$ does not have a bias term, we approximate $M_i^{-1}$ using its diagonal,
$P_i = \E[\texttt{diag}(xx^T)]^{-1}$. If $f_i$ has a bias term, 
we approximate $W_iW_i^T$ from \S \ref{sec:WW} with the incomplete Cholesky factorization
$P_i = \widetilde{W}_i\widetilde{W}_i^T$, where $\widetilde{W}_i$ approximates the respective $\Sigma_i$ term with its diagonal.
Note that this does \emph{not} approximate $M_i^{-1}$ with a diagonal matrix, since it still contains the
off-diagonal $\mu_i$ terms.

We construct the terms in $P_i$ using an EMA of the moments $\mu_i$ and $\chi_i = \E[x \odot x]$
across the minibatches\footnote{Note that these quantities are also straightforward to compute for convolutions:
$\mu_i$ is the mean over all pixels in the minibatch.}.
Then, $\E[\texttt{diag}(xx^T)] = \chi_i$ and
$\texttt{diag}(\Sigma_i)= \chi_i - \mu_i \odot \mu_i$.
When initializing conjugate gradient with the solution from the previous LNB step and using 
$P_i$ as the preconditioner, we found that we obtain diminishing returns
on the solution quality after only two iterations in practice. The online version of LNB is
summarized in Algorithm \ref{algo2}. Note that this approach is also amenable to the distributed
training setting where the sufficient statistics, $\mu$ and $\chi$, can be asynchronously updated and communicated
to different nodes (as with the gradient vectors) while each node runs its own conjugate gradient solver.

\subsection{Complexity}
The space complexity of computing a LNB step is the same as backpropagation since
it reuses the feature sets $X_B$. However, the actual amount of used memory is increased by
approximately the size of $X_B$ in order to evaluate the MVP. Additionally, conjugate
gradient holds additional copies of the parameters.

For each neuron, the time complexity to solve the linear system is $O(n_id_i^2 + d_i^3)$, in general.
However, in practice, due to the outer-product nature of the metric, one MVP evaluation is
$O(n_i d_i)$ and we perform only 2 steps of conjugate gradient.

\section{Related Work}
\label{sec:connections}
\subsection{Function Space Optimization and Whitening}
The functional gradient descent framework is general and the 
application to neural networks described in \citet{grub2010} has connections to
multiple recent works. The LocoProp-S Algorithm 1 in \citet{locoprop}
is mechanically equivalent to solving \Eqref{eq:OLS2} using gradient descent
instead of solving in closed form, whereas this work avoids constructing the targets altogether.
As the authors discuss in Appendix C,
there is a reduction from LocoProp-S to ProxProp \citep{proxprop}.

The FOOF algorithm \citep{foof} is also motivated by performing gradient descent
in function space. For simplicity in implementation the authors analyzed functions
with no bias terms and Equation 6 in \citet{foof} is equivalent to
solving \Eqref{eq:normal}. In that work, an online estimate of the 
covariance matrix is computed but only infrequently inverted to update the
preconditioner. This work provides the following novel perspectives. First, 
we formulate the metric via inner-product of Jacobians. This approach
automatically handles, without specialized implementations, various forms of the neurons,
including if it has a bias term, is a convolutional layer, etc. 
Second, we make the equivalence to feature whitening explicit, which enables 
us to analytically identify a Cholesky preconditioner for fast convergence via
conjugate gradient. Lastly, this feature whitening perspective also
explains why FOOF converges in a single step in the example described in Appendix F:
it is performing a Newton step on squared loss, which is equivalent to performing one
gradient descent step on whitened features (via the reparameterized model).

Standardizing the inputs to neurons is an essential practitioner technique \citep{LeCun2012},
with BatchNorm \citep{batchnorm} being one of the most prevalent
usages with a diagonal approximation. The reparameterized model interpretation discussed in
\S \ref{sec:WW} is equivalent to placing a BatchNorm layer before applying the \emph{weights},
as opposed to typical practice of placing BatchNorm before the
\emph{activation}. The preconditioner connection is also discussed in \citep{lange}
where the preconditioner for CNNs is carefully constructed.
Related, PRONG \cite{nnn} also computes the same first and
second moments (again, manually) to reparameterize the network, but misses the analytical interpretation
of the preconditioner matrix that enables easy adaption to arbitrary network topologies.

In summary, all aforementioned work will obtain the same result, and the primary difference
is implementation. The primary benefit of the LNB interpretation is that it can be easily
implemented for any differentiable network architecture and the trust-region connection
informs an adaptive step size.

\subsection{(Quasi-) Second-order methods}
The $M_i$ metric is very local in that it finds
an update vector that induces an $\eps$-sized squared norm in the differential of each neuron's output,
$\langle \delta\theta_i, \delta\theta_i \rangle_{M_i} =
\delta\theta_i^T \E[\frac{\partial x_i}{\partial \theta_i}^T\frac{\partial x_i}{\partial \theta_i}] \delta\theta_i
= \E[\delta x_i^T \delta x_i]$.
Intuitively, a better metric would compose the neuron's output differential with the change in the \emph{network's}
output differential, $\delta x_m$, to ensure that the cascaded output perturbations are small when the neuron's parameters
change; this corresponds to the Gauss-Newton (GN) matrix,
$\E[\frac{\partial x_i}{\partial \theta_i}^T \frac{\partial x_m}{\partial x_i}^T\frac{\partial x_m}{\partial x_i} \frac{\partial x_i}{\partial \theta_i}]
= \E[\frac{\partial x_m}{\partial \theta_i}^T\frac{\partial x_m}{\partial \theta_i}]$.
A seemingly better metric would measure how $\delta x_m$ locally varies in the landscape
of the loss function, i.e., the inner product of $\delta x_m$ along the eigenvectors of the 
loss' Hessian at $x_m$,
$\E[\frac{\partial x_m}{\partial \theta_i}^T \frac{\partial^2 \ell_y(x_m)}{\partial x_m \partial x_m}  \frac{\partial x_m}{\partial \theta_i}]$;
this is the Generalized GN matrix (GGN) \citep{Schraudolph2002}.
As reviewed in \citet{kunster2019},
in the special case when the network predictions, $x_m$, are the parameters of an exponential probability
distribution and $\ell(x_m)$ is log-loss, then the GGN matrix is equivalent to the FIM used in
natural gradient descent.

There are strong theoretical reasons to prefer the FIM in general
\citep{amari1998,martens2020}; however,
computing the metric is typically expensive in practice and there is
extensive research for efficiently approximating it
\citep{martens2015,Ren2021TensorNT}.
The functional gradient descent interpretation makes no attempt to approximate the FIM.
But, given that LNB uses a much simpler metric, it
provides an effective and efficient intermediary between the identity metric of gradient descent
and the full network Jacobian used in GGN and natural gradients.

\begin{figure}[t]
    \begin{center}
    \centerline{\includegraphics[width=0.5\columnwidth]{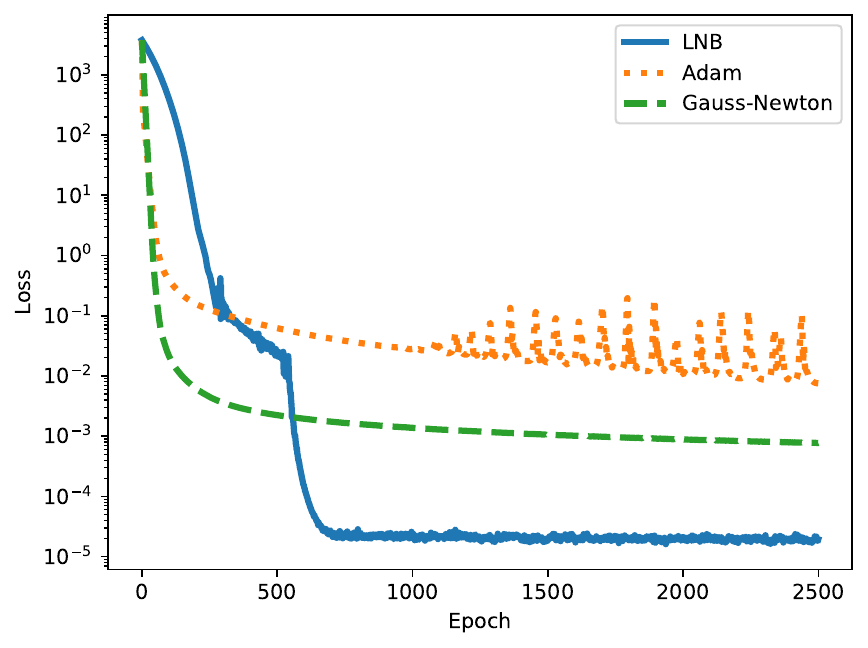}}
    \caption{Matrix factorization via a 2-layer linear network}
    \label{recht}
\end{center}
\end{figure}

\section{Experiments}
\label{sec:experiments}
The prior works of FOOF, LocoProp, PRONG have shown to compare competitively
with other sophisticated optimizers such as K-FAC \citep{martens2015}.
Given their discussed equivalence with LNB, we focus on experimentally confirming
the contributions of this work: feature whitening via preconditioning the
gradient vector, and the applicability and effectiveness on
the realistic networks of ViT \citep{vit} and UNet \citep{unet}.

Due to its de facto status, we benchmark convergence with Adam in both
iteration and wall time.
In all experiments, the default EMA values were used ($\beta_1=0.9$, $\beta_2=0.999$)
for Adam while the learning rate was grid searched around $1e^{-4}$
All timings were recorded from a NVIDIA L4 GPU. Due to the deterministic nature of performing
2 conjugate gradient steps for LNB, the variance in reported times is negligible.

\subsection{Matrix factorization}
We start with the pathological example from \citet{recht}. This is a matrix
factorization problem formulated as a two-layer linear network:
$\sum_{i=1}^{n} \Vert W_1 W_2 x_i - y_i\Vert^2$,
where $y_i=Ax_i$ for a poorly conditioned matrix, $\kappa(A)=10^5$. Due to the conditioning
and the columns being correlated, it is known that gradient descent converges slowly for this
problem, whereas GN converges quickly. Because the LNB preconditioner is decorrelating
the feature space, we would also expect fast convergence.

We use the same initialization as in the notebook, but in order to magnify the differences,
we increase the dimensions by a factor of $10$: $n=10^4$,
$W_2 \in \sR^{60 \times 60}$, $W_1 \in \sR^{100 \times 60}$. Learning rates were tuned via
grid search to find fast and stable convergence for each method. The results are plotted in Figure \ref{recht}
and reproduce the prior reported slow convergence of Adam and demonstrate fast convergence with LNB.

\begin{figure}[t]
    \begin{center}
    \subfigure[Original pixels ($x$)]{\includegraphics[width=0.49\columnwidth]{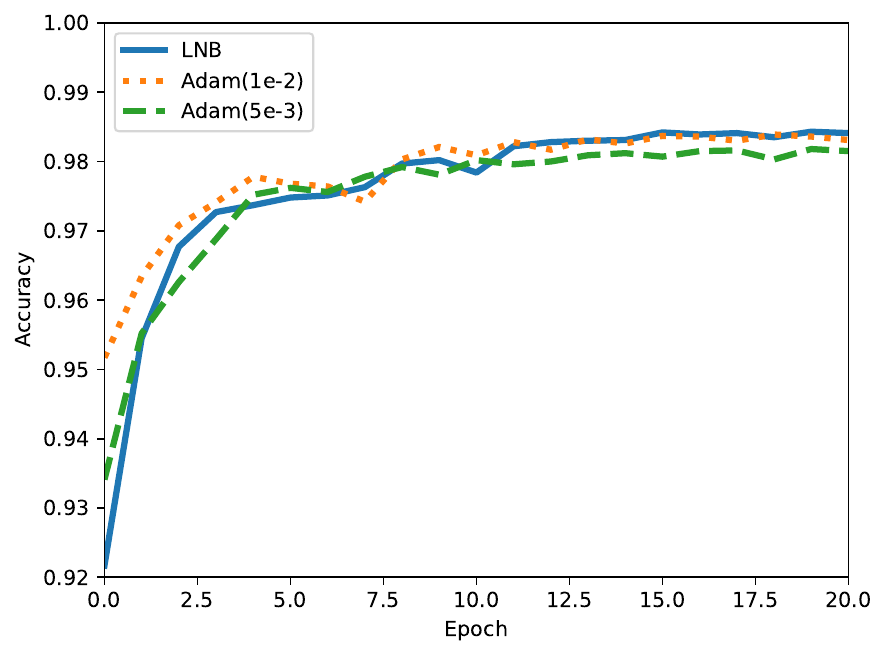}}
    \subfigure[Inverted pixels ($1-x$)]{\includegraphics[width=0.49\columnwidth]{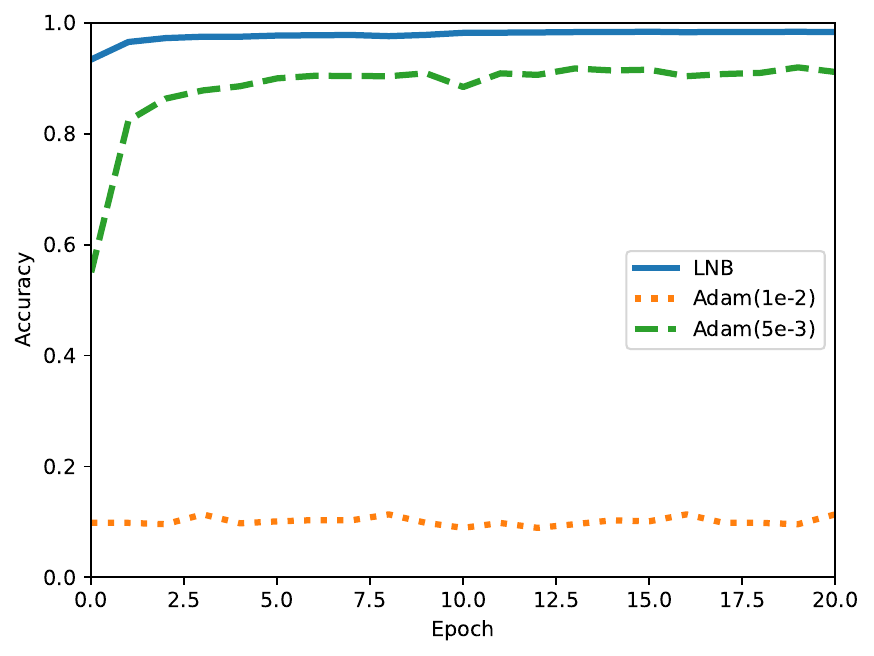}}
    \vskip -0.1in
    \caption{MNIST test accuracy evolution trained on the original data (a) vs. inverted pixels (b).
    The step-size is parenthesized.}
    \label{fig:mnist}
    \end{center}
\end{figure}

\subsection{MLP}
We reproduce the MLP result in \citet{grub2010} that compares boosting and gradient
descent for a 2-layer MLP on MNIST using 800-node layers, $\tanh$ activation,
Glorot Normal initialization and a batch size of 1,000.

In Fig \ref{fig:mnist}-a., we plot the test accuracies w.r.t. epoch with the best two learning rates for Adam.
We first note that Adam and LNB obtain better than the prior reported accuracy of $98.3\%$.
Second, LNB achieved this performance with a fixed (ridge) regularizer $\lambda$, whereas prior
work heavily tuned this.
One explanation for the difference is that LNB is taking an adaptive step size according to the metric
and this was not derived before.
Third, there is little observed difference between boosting and gradient descent on this dataset.
This can be explained due to that most of the binary pixels in MNIST are zero, so the feature space
of the vectorized images is low rank, i.e., decorrelating provides little benefit.

\begin{figure}[t]
    \begin{center}
    \subfigure[\texttt{train} loss]{\includegraphics[width=0.49\columnwidth]{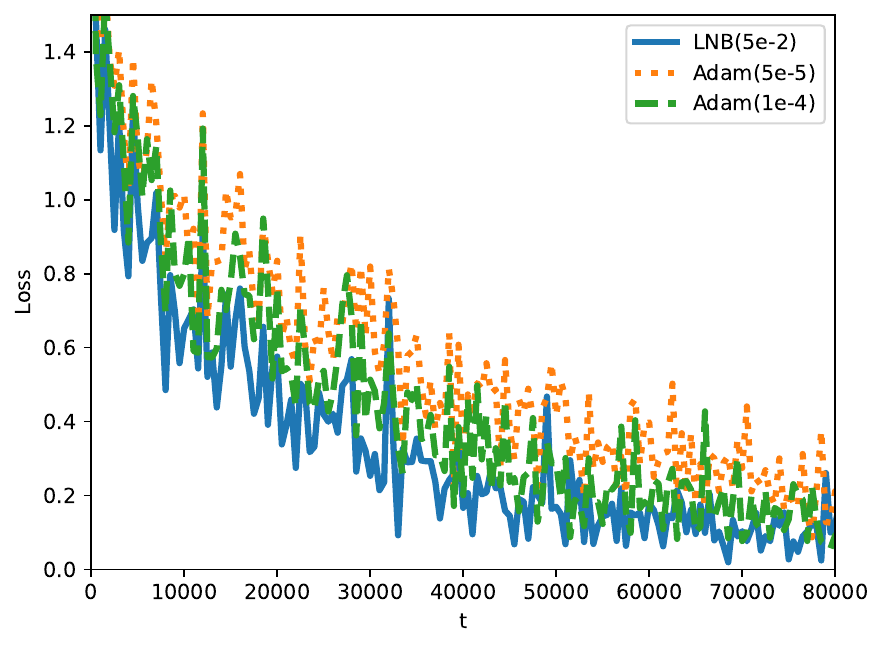}}
    \subfigure[\texttt{test} accuracy]{\includegraphics[width=0.49\columnwidth]{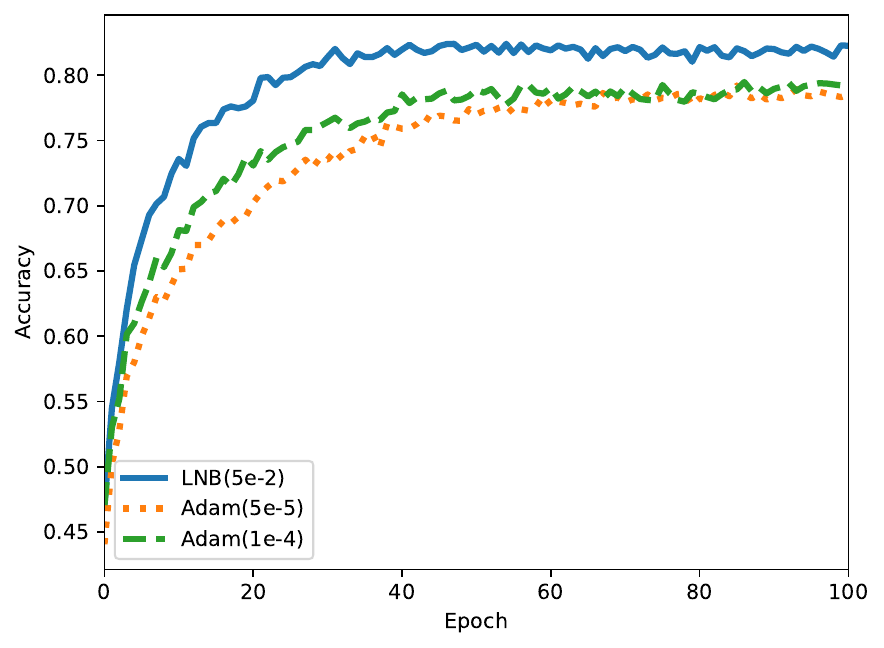}}
    \caption{ViT performances on CIFAR10.}
    \label{fig:vit}
    \end{center}
\end{figure}

However, whitening does include a centering step. If we were to shift the feature space, we would expect
to get same performance. In Fig \ref{fig:mnist}-b, we plot the same models when trained and tested on pixels
$1-x$, where $x$ are the original binary pixel values used in Fig \ref{fig:mnist}-a. We observe that LNB gets very similar performance
while Adam (and other methods not invariant to affine reparameterizations) degrade. Although we could (and should)
simply normalize the features before training, this example illustrates a case of how feature scaling can
greatly affect convergence.

\subsection{Vision Transformer}
We train a vision transformer \cite{vit} using the notebook from Equinox \cite{eqx}
on CIFAR10.
The only modification we make is to not learn the affine terms in the LayerNorm
in order to speed up experimentation and we observed no performance benefit with it.
The train and test fold performances are show in \Figref{fig:vit}, where we observe
faster convergence and better generalization with LNB. Excluding JIT compilation time,
the duration per epoch for LNB and Adam is 1.26 min and 0.85 min, respectively. 

\begin{figure}[t]
    \begin{center}
    \subfigure[\texttt{train} loss]{\includegraphics[width=0.49\columnwidth]{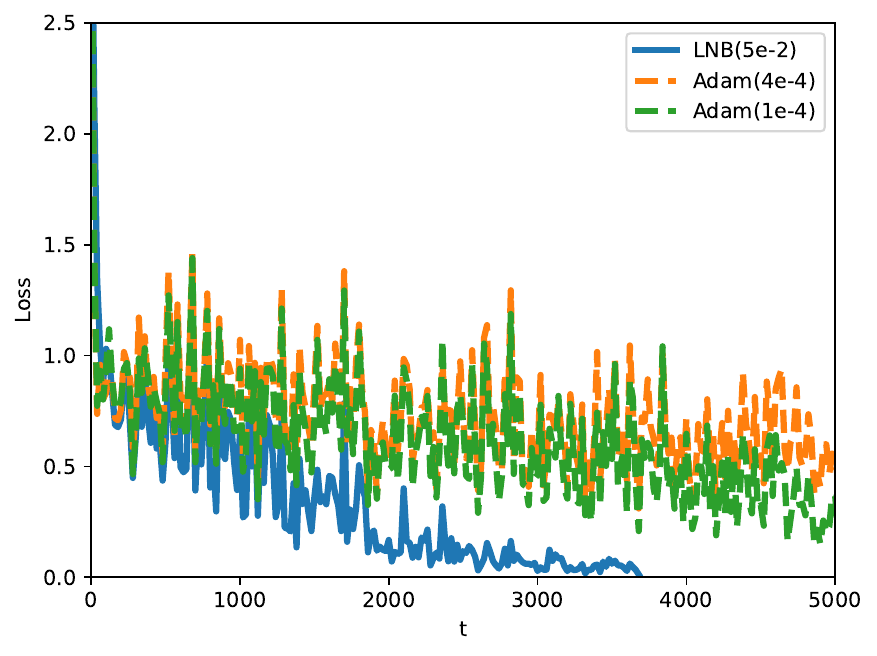}}
    \subfigure[\texttt{val} accuracy]{\includegraphics[width=0.49\columnwidth]{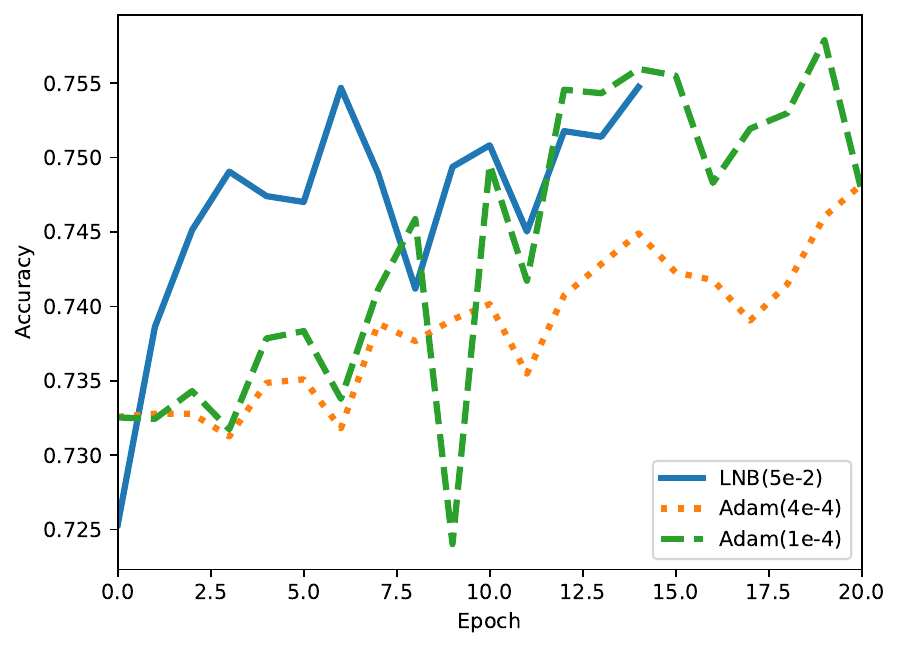}}
    \caption{UNet performances on VOC Segmentation.}
    \label{fig:unet}
    \end{center}
\end{figure}

\subsection{UNet}
We train a UNet \cite{unet} on the 2012 VOC Segmentation Challenge dataset \cite{voc}. The images are
pixelwise normalized into the range $[0,1]$ using ImageNet mean and variance R,G,B pixel values
and then zero-padded to $500 \times 500$ size and then downsized to $384 \times 384$.
No data augmentation is performed. We plot the results in \Figref{fig:unet} and
remark that LNB converges very quickly, using the same learning rate as with ViT, and
avoids overfitting. While both optimizers converge to comparable performance on the
\texttt{val} split, the rapid progress by LNB suggests it would be able to leverage
more data effectively.
However, excluding JIT compilation time,
the duration per epoch for LNB and Adam is 2.92 min and 1.27 min, respectively, and
this highlights the trade-off between convergence w.r.t. iterations vs. wall time.
While LNB converged marginally faster in wall time and is significantly
easier to implement to prior equivalent work, it is future work to better understand
in what deep networks does the whitening behavior lead to better generalization as
demonstrated in the other three experiments.

\bibliography{lnb}

\end{document}